\documentclass{article}
\usepackage{spconf,amsmath,graphicx,latexsym, amssymb, subcaption}

\usepackage{enumitem}
\setlist{nosep, leftmargin=14pt}

\usepackage{mwe} % to get dummy images
\usepackage{url}
\usepackage{xcolor}
\usepackage{afterpage}
\usepackage{float}
\usepackage{booktabs}
\usepackage{stmaryrd}

\title{GLFNet: Global-Local (Frequency) Filter Networks for efficient Medical Image Segmentation}

\name{
    \begin{tabular}{c}
        Athanasios Tragakis$^{1\star}$, Qianying Liu$^{2\star}$, Chaitanya Kaul$^{2}$, Swalpa Kumar Roy$^{3}$ \\
        Hang Dai$^{2}$, Fani Deligianni$^{2}$, Roderick Murray-Smith$^{2}$, Daniele Faccio$^{1}$
    \end{tabular}
}

 \address{
 $^{1}$ School of Physics and Astronomy,University of Glasgow, Glasgow G12 8QQ, UK \\
 $^{2}$ School of Computing Science, University of Glasgow, United Kingdom, G12 8RZ \\
 $^{3}$ Department of Computer Science and Engineering, Alipurduar Government \\
 Engineering and Management College, West Bengal, 736206, India}   

\begin{document}
%\ninept
%

\maketitle
\begin{abstract}
We propose a novel \textit{transformer-style} architecture called Global-Local Filter Network (GLFNet) for medical image segmentation and demonstrate its state-of-the-art performance. We replace the self-attention mechanism with a combination of global-local filter blocks to optimize model efficiency. The global filters extract features from the whole feature map whereas the local filters are being adaptively created as $4 \times 4$ patches of the same feature map and add restricted scale information. In particular, the feature extraction takes place in the frequency domain rather than the commonly used spatial (image) domain to facilitate faster computations. The fusion of information from both spatial and frequency spaces creates an efficient model with regards to complexity, required data and performance. We test GLFNet on three benchmark datasets achieving state-of-the-art performance on all of them while being almost twice as efficient in terms of GFLOP operations. Our code is available here$^{\dagger}$.
\def\thefootnote{$\star$}\footnotetext{Equal Contribution}
\def\thefootnote{$\dagger$}\footnotetext{\url{https://github.com/Thanos-DB/GlobalLocalFilterNetworks}}
\end{abstract}

\section{Introduction}
\label{sec:intro}
With the development of deep learning for computer vision tasks, convolutional neural networks (CNNs) were the first choice for a very long time. This choice also dominated the approaches used in the field of medical imaging segmentation, a sub-field of computer vision with applications in automated diagnosis. Currently, UNet \cite{Ronneberger2015} and its variants \cite{Huang2020, kaul2021focusnet++} with CNN backbones dominate medical image segmentation, as they model local attributes inside their receptive fields.\\
Since the Vision Transformer \cite{vit}, the field has seen many advancements due to its ability to capture global dependencies. However, alongside performance gains, there has also been an increase in complexity of such models. This results in a need for less complex and more efficient transformers with high performance. The Swin Transformer \cite{cao2022swin} achieved better performance by using self-attention on smaller windows. This reduces the global self-attention's quadratic complexity to linear \cite{vit}. The \textit{Fast Attention Via positive Orthogonal Random} technique for approximating softmax and Gaussian kernels \cite{performers} also showed linear complexity. GFNet \cite{gfnet} replaces the self-attention with a series of operations that include the Fast Fourier Transform (FFT) algorithm reducing the quadratic complexity to a log-linear one. This way no inductive bias is introduced to the learning process, improving the generalization ability of the model but at the cost of data efficiency. Similarly, Multi-Layer Perceptron (MLP) models and most Transformer models also share the same traits and weaknesses. Larger datasets are needed to achieve competitive performance. This leads to either having to pre-train on huge domain relevant datasets, or gather new ones. Both solutions introduce further problems such as increased costs, increased training times, higher training complexity due to parallelization and larger carbon footprint. However, medical imaging datasets are often sparse, gathering more data is difficult and pre-training on out-of-distribution datasets has an improvement ceiling. Most existing Transformers \cite{transunet, cao2022swin, valanarasu2021,Lin2022} for medical segmentation use off-the-shelf Transformer blocks, which lack inductive bias and therefore have difficulty generalizing on small datasets, limiting the performance.

Keeping in mind the need for more efficient architectures in medical imaging, we propose GLFNet, a \textit{transformer-style} architecture specifically designed for medical image segmentation, able to alleviate current Transformer's drawbacks and surpass the performance of previous state-of-the-art architectures by combining global and local information. The global filters are learned in a similar way as GFNet \cite{gfnet} by first converting the feature maps to the frequency domain using a 2D Fourier transform, estimating a global filter kernel in the frequency domain and then performing the inverse 2D Fourier transform to revert back to the spatial domain. Alongside the global filters, local filters are used to capture more precise organ boundaries by introducing inductive biases which alleviate the data-hungry properties of Transformers. Approximating convolutions in the frequency domain makes these models faster and at the same time, more data efficient. Our model is \textit{transformer-style} as it keeps the general structure of the transformer block, while replacing self attention with frequency based weight kernel estimations. Our main contributions are two-fold:
\begin{itemize}
    \item We propose a novel transformer-style block specifically constructed for medical imaging which can replace current transformer layers. It fuses global and local information from the frequency domain and shows favorable characteristics such as complexity and data efficiency.
    \item We propose our Global-Local Filter Network (GLFNet) model based on the new transformer-style block, that surpasses all baselines for the task of medical image segmentation on multiple datasets and modalities.
    % \item We show, through ablation studies the importance of the novel parts of the model and the impact they have on performance.
\end{itemize}

\section{GLFNet}
\label{sec:glfnet}

Given a dataset $\{\textbf{X},\textbf{Y}\}$, where $\textbf{X}$ are the input images for the model and $\textbf{Y}$ the ground truth segmentation masks, the model learns to predict a segmentation mask $\hat{\textbf{Y}}$. The input image is a 2D slice of a 3D volume with the exception of BraTS19 which is a multi-modal MRI dataset. Here we concatenate the data together to create a four-channel input and perform an early fusion of the four modalities. This shows the flexibility of the model to adapt to multi-modal and single-modal datasets without making any architectural changes. 

The model follows the Fully Convolutional Transformer (FCT), a U-Net like architecture firstly introduced by \cite{fct} which is the first pure 2D Transformer in medical image segmentation. We replace the FCT attention block with our GLFNet block while retaining the Wide-Focus module as a replacement to MLPs to introduce further inductive biases through multi-scale feature processing.

\afterpage{%
  \begin{figure*}
  \centering
  \includegraphics[trim={5.5cm 8.3cm 9.cm 7.5cm}, clip=true, scale=1.]{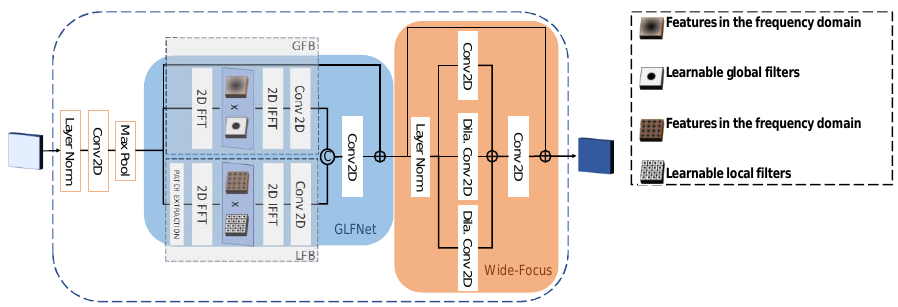}
  \caption{ The Global-Local Filter Network block. Each input is passed through a layer normalization, two convolutional layers and a max pooling operation before the GLFNet module. We use the max pooling operation before the GLFNet block to reduce computations. The GLFNet has one branch for global filters, one for local filters and lastly a skip connection. The output of the GLFNet module is fed to the Wide-Focus module. GFB stands for Global Filter Branch and LFB for Local Filter Branch. }
  \label{model}
  \end{figure*}
}

%\begin{table}[H]
%\centering
%\resizebox{\columnwidth}{!}{%
%\begin{tabular}{p{2cm}||p{5cm}}
%\textbf{Method} & \textbf{Complexity (FLOPs)} \\
%\hline
%Convolution & $\mathcal{O}(\textcolor{red}{C_{ik}^2HWC_{o}})$ \\
%\hline
%Depthwise Convolution & $\mathcal{O}(Ck^2HW)$ \\
%\hline
%Self-Attention & $\mathcal{O}(C^2HW+C(HW)^2)$ \\
%\hline
%MLP & $\mathcal{O}(C(HW)^2)$ \\
%\hline
%GFNet & $\mathcal{O}(CHW\lceil\log_2(HW)\rceil+CHW)$ \\
%\hline
%\hline
%GLFNet & $\mathcal{O}(CHW\lceil\log_2(HW)\rceil+CHW)$ \\
%\end{tabular}
%}
%\caption{The complexities of the most well known operations in computer vision. C represents the number of channels, H the height, W the width and k the kernel size in convolutional layers. GLFNet has the same complexity of GFNet due to the FFT operations but also is more data efficient.}
%\label{acdc}
%\end{table}

\subsection{The GLFNet Block}

\begin{figure}[t]
    \centering
    \begin{minipage}[t]{8.5cm} % Adjust the width as needed
        \centering
        \includegraphics[trim={8cm 5.5cm 7.5cm 3cm}, scale=.5, clip=true]{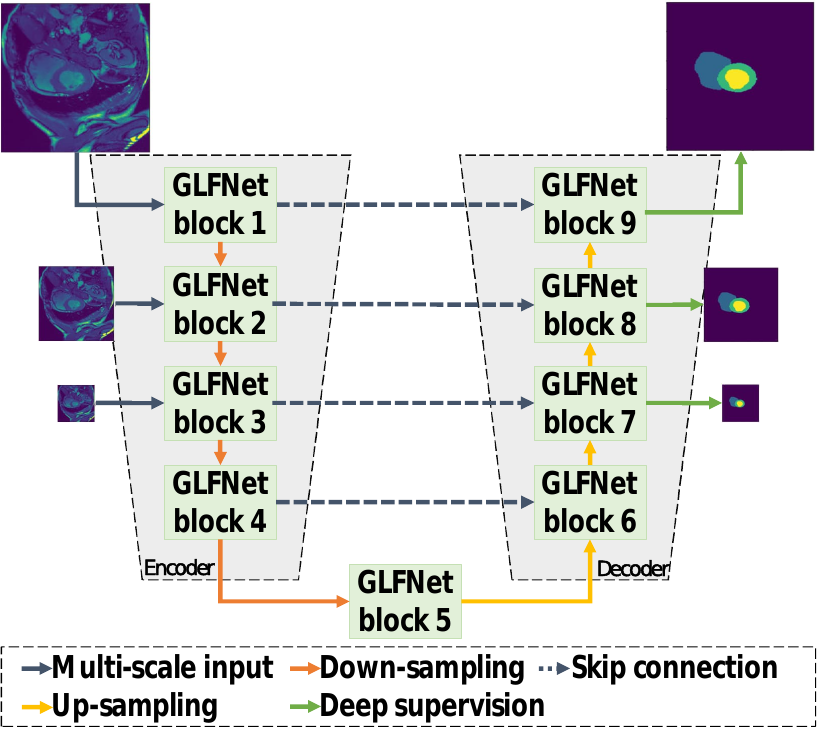}
        \caption{The proposed architecture is a UNet-like architecture with the difference that each block is a GLFNet block. To boost performance we utilize a multi-scale input for scale invariance as well as deep supervision . The encoder and decoder are symmetric.}
        \label{overview}
    \end{minipage}
\end{figure}

Each GLFNet block starts with a normalization layer, followed by two convolutional layers. The extracted features are then down-sampled by a max pooling operation. The max pooling output is copied and spilt into two parallel branches to be processed by local and global filters. Their outputs are then fused via concatenation and is followed by a convolution to extract fused features. The global branch performs a 2D Fourier transformation of the feature maps to convert them to the frequency domain, followed by an element-wise multiplication with learnable filters, after which the inverse Fourier transformation reverts the features back to the spatial domain. Parallelly, the local branch first extracts $4\times4$ patches from the slice features, and then applies the same series of operations locally for each patch (separately) to restrict the interactions to within each patch. This way backpropagation learns to approximate locality by being jointly trained with the global filter. Since the two branches learn spatial interactions in the frequency domain while all filters operating across all frequencies, both branches can capture short and long term spatial dependencies in the data. %Although the global branch will be able to generalize better, the local branch has higher data efficiency by having locality as prior knowledge. 

The fused features obtained via the convolutional layer are added via a residual connection to the input of the branches to eliminate any vanishing gradient problems and ensure smoother convergence. Finally, the Wide-Focus module extracts multi-scale features by dilating convolutions at multiple dilation rates. By using Wide-Focus and not MLPs the spatial context in the features is better preserved without the need to flatten the feature maps. Wide-Focus makes use of a multi-branch residual block of multi-dilated convolutional layers to increase the receptive fields and achieve a better contextual understanding of the features.

\subsection{Encoder--Decoder}

Our architecture follows the familiar encoder-decoder structure. For the multi-modal MRI dataset we use only one encoder, rather than one per modality, which forces it to extract shared features among all 4 modalities after an early fusion. As such the architecture of the model is not changed based on the modality or the number of modalities. 

The encoder is comprised of 4 GLFNet blocks followed by another GLFNet block for the bottleneck. We make use of a multi-scale input to improve generalization and scale invariance. As our network is symmetric in structure, the decoder also comprises of 4 GLFNet blocks, with the difference that they are adapted to upsample the feature maps to reach the dimensions of the input images for the final prediction of the segmentation masks. Finally, we use deep supervision, a technique that motivates blocks that are placed before the final output to approximate intermediate predicted masks to further improve performance.

\section{Experiments}
\label{sec:experiments}
We showcase the performance of our model by training and evaluating it on 3 datasets with 2 different modalities and one being multi-modal. We train on the Automatic Cardiac Diagnosis (ACDC) \cite{acdc} (MRI), Synapse Multi-organ Segmentation Challenge$^{1}$ (CT) and Multimodal Brain Tumor Segmentation Challenge 2019 (BraTS19) \cite{bratsA, bratsB, bratsC}. The evaluation metric is the Dice coefficient \cite{diceA, diceB}.

\textbf{Datasets.} The ACDC dataset contains 100 MRI scans in total for left ventricle (LV), right ventricle (RV) and myocardimum (MYO). We follow the same split as FCT. The Synapse dataset contains 30 scans and we follow the same split as \cite{transunet} to predict masks for classes aorta, gallbladder, left kidney, right kidney, liver, pancreas, spleen and stomach. For the BraTS19 dataset, we use both high-grade gliomas and low-grade gliomas data with four modalities. 80$\%$ of the data is used for training and 20$\%$ for validation. The final evaluation areas are whole tumor (WT), tumor core (TC) and enchancing tumor (ET).

\textbf{Implementation Details.} All our models are trained from scratch on one RTX3090 using Tensorflow 2.5.0. %The baseline for ACDC and Synapse is FCT and for Brats19 CS-Unet as those are closer to the proposed architecture making it a fairer comparison.

\subsection{Comparison with Existing Methods} 

\def\thefootnote{$1$}\footnotetext{\url{https://www.synapse.org/#!Synapse:syn3193805/wiki/217789}}

The results and visualizations on three datasets are summarized through Table \ref{acdc}, Table \ref{synapse_results}, Table \ref{brats} and Figure \ref{vis}. GLFNet outperforms all existing baselines across the ACDC and BraTS19 datasets. For the Synapse  dataset, our model gets the best average dice score across all baselines, and the best performance on 5 of the 8 organ classes showcasing our models' ability to delineate organs well. GLFNet (18.7G FLOPS) outperforms the previous state-of-the-art FCT (28.7G FLOPS) in terms of both performance, as well as efficiency when both models contain similar parameters.

\begin{table}[H]
\centering
\resizebox{\columnwidth}{!}{%
\begin{tabular}{p{2.5cm}|| p{1cm}| p{1cm}| p{1cm}| p{1cm}}
\toprule
{Method} & Avg. & RV & MYO & LV\\
\hline
TransUNet \cite{transunet} & 89.71 & 88.86 & 84.53 & 95.73\\
\hline
Swin UNet \cite{cao2022swin} & 90.00 & 88.55 & 85.62 & 95.83\\
\hline
CS-Unet \cite{csunet} &  91.37  &  89.20 &  89.47 & 95.42 \\
\hline
nnUNet \cite{nnunet} & 91.61 & 90.24 & 89.24 & 95.36\\
\hline
FCT \cite{fct} & 93.02 & 92.64 & 90.51 & 95.90\\
\hline
\hline
GLFNet & $\textbf{93.12}$ & $\textbf{92.69}$ & $\textbf{90.71}$ & $\textbf{95.97}$\\
\bottomrule
\multicolumn{5}{r}{\scriptsize \textbf{Best} is reported as bold \hfill}
\end{tabular}
}
\caption{Segmentation results on ACDC dataset.}
\label{acdc}
\end{table}

\begin{table}[H]
\centering
\resizebox{\columnwidth}{!}{%
\begin{tabular}{p{2.5cm}|| p{1cm}| p{1cm}| p{1cm}| p{1cm}}
\toprule
{Method} & Avg. & WT & TC & ET\\
\hline
Swin Unet \cite{cao2022swin} & 81.45 & 88.72 & 77.89 & 77.74\\
\hline
TransUNet \cite{transunet} & 82.31 & 88.26 & 80.57 & 78.09\\
\hline
CS-Unet \cite{csunet} & 83.36 & 89.03 & 81.19 & 79.86\\
\hline
\hline
GLFNet & $\textbf{85.51}$ & $\textbf{90.55}$ & $\textbf{84.38}$ & $\textbf{81.59}$\\
\bottomrule
\multicolumn{5}{r}{\scriptsize \textbf{Best} is reported as bold \hfill}
\end{tabular}
}
\caption{Segmentation results on BraTS19 dataset.}
\label{brats}
\end{table}

\begin{table*}[ht]
\begin{center}
\begin{tabular}{p{3cm}|| p{1cm}| p{1cm}| p{1cm}| p{1.2cm}| p{1.2cm}| p{1cm}| p{1cm}| p{1cm}| p{1cm}}
\toprule
{Method} & Avg. & Aorta & GB & Kid. (L) & Kid. (R) & Liver & Panc. & Spl. & Stom. \\
\midrule
TransUNet \cite{transunet} & 77.48 & 87.23 & 63.13 & 81.87 & 77.02 & 94.08 & 55.86 & 85.08 & 75.62 \\
\hline
Swin UNet \cite{cao2022swin} & 79.13 & 85.47 & 66.53 & 83.28 & 79.61 & 94.29 & 56.58 & 90.66 & 76.60 \\
\hline
CS-Unet \cite{csunet} &  82.21 & 88.40& 72.59	&  85.28& 	79.5& 	94.35& 	{\bf 70.12}& 91.06& 	 75.72 \\
\hline
FCT \cite{fct} & 83.53 & \textbf{89.85} & 72.73 & 88.45 & 86.60 & \textbf{95.62} & 66.25 & 89.77 & 79.42 \\
\hline
\hline
GLFNet & \textbf{85.88} & 89.73 & \textbf{77.10} & \textbf{93.49} & \textbf{90.83} & 95.58 & 66.33 & \textbf{91.98} & \textbf{82.01} \\
\bottomrule
\multicolumn{10}{r}{\scriptsize \textbf{Best} is reported as bold \hfill}
\end{tabular}
\end{center}
\caption{Segmentation results on Synapse. GB is Gallbladder, Kid. Kidney, Panc. Pancreas, Spl. Spleen and Stom. Stomach.}
\label{synapse_results}
\end{table*}

\afterpage{%
    \begin{figure}[t]
  \centering
  \includegraphics[width=1\linewidth]{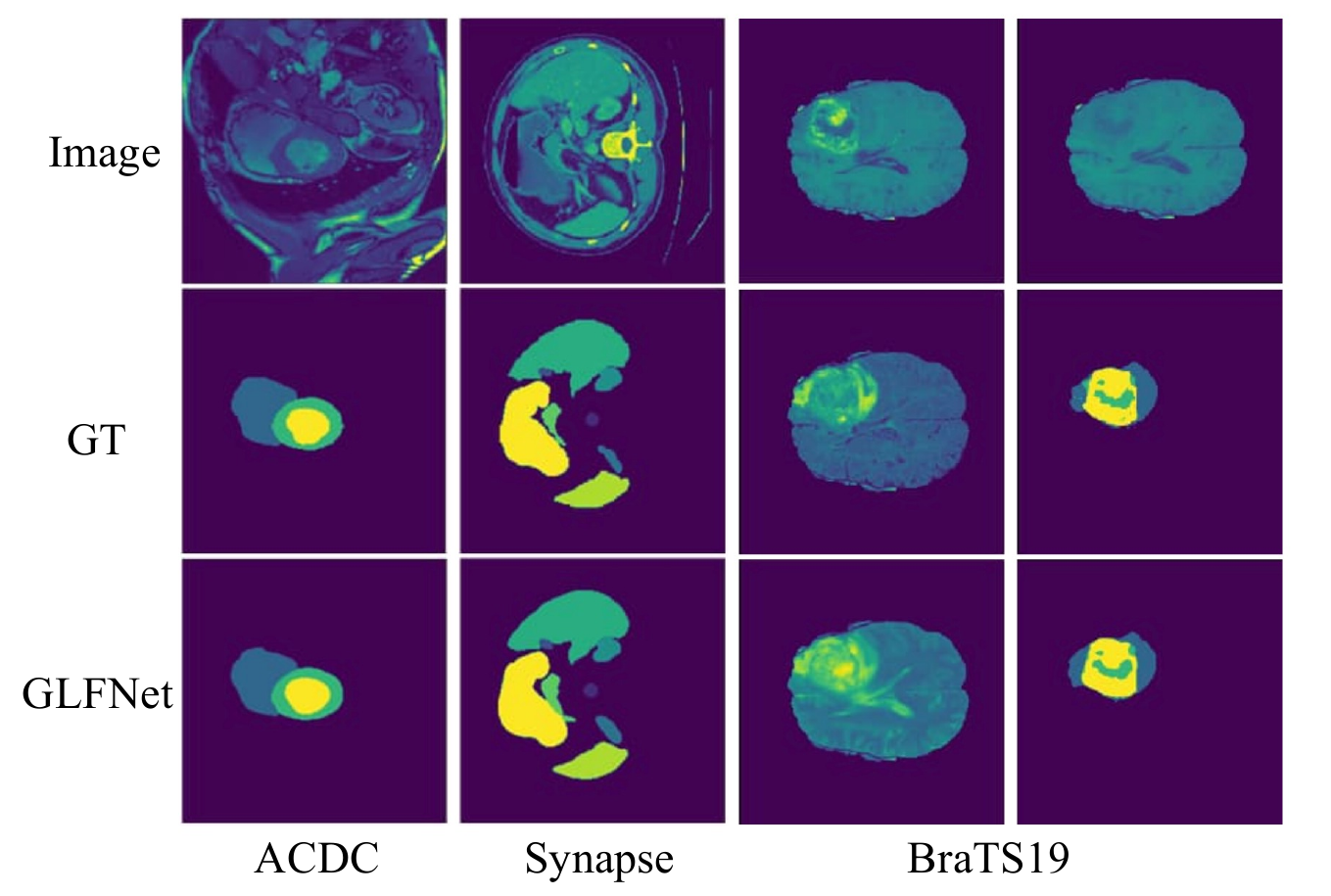}
  \caption{ Qualitative results of GT and the GLFNet prediction.}
  \label{vis}
  \end{figure}
}

\subsection{Ablation Study}
\label{sec:ablation}

We perform ablations on the ACDC dataset specifically to demonstrate the effectiveness of having a local and global filter branch in parallel in our model. Table \ref{ablation} shows that extracting local and global frequency features jointly outperforms having just either one extracted by the model.

\begin{table}[H]
\centering
\resizebox{\columnwidth}{!}{%
\begin{tabular}{p{2.5cm}|| p{2.5cm}| p{2.5cm}}
\toprule
Branch A & Branch B & Avg.\\
\hline
GFB & -- & 92.88 \\
\hline
LFB & -- & 92.82 \\
\hline
GFB & GFB & 93.01 \\
\hline
LFB & LFB & 92.99 \\
\hline
GFB & LFB &  \textbf{93.12} \\
\bottomrule
\end{tabular}
}
\caption{Demonstrating the importance of the proposed dual branch GLFNet with both global and local filter blocks.}
\label{ablation}
\end{table}

\section{Conclusion}
\label{sec:conclusion}
In this paper, we proposed GLFNet, which extracts global and local features from medical images in the frequency domain to perform accurate and efficient medical image segmentation. GLFNet learns a combined feature space of local and global feature relations to provide an inductive bias towards local neighbourhoods of image regions. This translates to highly accurate segmentation results, as demonstrated on three benchmark datasets. GLFNet achieves this performance with half the number of GFLOPS compared to the state-of-the-art Fully Convolutional Transformer for the same number of model parameters. Given that the computational complexity of the Fourier transform is log-linear, GLFNet reduces the complexity of \textit{transformer-style} architectures drastically from quadratic without a reduction in performance.

%\vfill
%\pagebreak

\section{Compliance with Ethical Standards}
This research study was conducted retrospectively using human subject data made available in open access. Ethical approval was not required as confirmed by the license attached with the open access data

\section{Acknowledgements}
D.F., A.T. acknowledge support from Royal Academy of Engineering through the Chairs in Emerging Technologies scheme. Q.L., F.D. acknowledge funding from China Scholarship Council and funding from grant EP/W01212X/1. D.F., R.M.S., C.K. acknowledge funding from \textit{QuantIC} Project funded by EPSRC Quantum Technology Programme (grant EP/MO1326X/1, EP/T00097X/1), and \textit{Google}. R.M.S, C.K. acknowledge EP/R018634/1, and R.M.S acknowledges EP/T021020/1. For the purpose of open access, the author(s) has applied a Creative Commons Attribution (CC BY) license to any Accepted Manuscript version arising.


\bibliographystyle{IEEEbib}
\bibliography{refs}
\end{document}